\documentclass[11pt]{article}

\usepackage[final]{acl}
\usepackage{times}
\usepackage{latexsym}
\usepackage[T1]{fontenc}
\usepackage[utf8]{inputenc}
\usepackage{microtype}
\usepackage{inconsolata}
\usepackage{booktabs}
\usepackage{amsmath}
\usepackage{graphicx}
\usepackage{xurl}
\usepackage{placeins}
\usepackage{todonotes}

\title{Do Cantonese-Adapted Language Models Better Predict Cantonese\\
Reading? A Cross-Model Eye-Tracking Evaluation}

\author{
  Ziqi Zhang$^{1}$ \quad
  Emmanuele Chersoni$^{1,2}$ \quad
  Mohammad Momenian$^{3}$
  \\[0.5em]
  $^{1}$Department of Language Science and Technology, The Hong Kong Polytechnic University\\
  $^{2}$Division of Artificial Intelligence and the Humanities, The Hong Kong Polytechnic University\\
  $^{3}$Department of Linguistics, The University of Hong Kong\\
  \texttt{\{zhangziqi.bram, emmanuelechersoni\}@gmail.com, momenian@hku.hk}
}

\begin{document}
\maketitle

\begin{abstract}
Information-theoretic measures derived from autoregressive language models are widely used to characterize the expectations that shape human reading, but whether language-variety-specific training improves such psycholinguistic alignment remains unclear.

This question is still open for Cantonese, where recent NLP evaluations reported mixed benefits from Cantonese-specific training relative to Mandarin-oriented or general-purpose models. 

Using naturalistic Cantonese eye-tracking data, we compare two within-family adaptation contrasts: CKIP GPT-2 Tiny versus its lightly Cantonese-adapted JED351 derivative, and Qwen2.5-7B versus CantoneseLLM-7B, which underwent substantially more extensive Cantonese continued pretraining and instruction tuning. From each model, we derive lexical surprisal, POS surprisal, entropy before the target, and entropy reduction. Lexical surprisal and the joint four-metric model consistently favor CantoneseLLM-7B, followed by Qwen2.5-7B, CKIP, and JED351, whereas entropy reduction favors CKIP. These results suggest that more extensive Cantonese-specific training can be associated with stronger predictive fit, while model rankings also depend on the information-theoretic measure being evaluated.
\end{abstract}

\section{Introduction}

Readers allocate different amounts of time to different words, making eye tracking a sensitive measure of online language processing \citep{attardo2023eye}. A central account of this variation is that processing difficulty increases when a word is unexpected in its context \citep{hale2001probabilistic,levy2008expectation}. Autoregressive language models operationalize this account by assigning a probability distribution to possible continuations. Their probabilities support not only lexical surprisal, but also measures of uncertainty before a word and probabilistic updating after it is observed \citep{hale2016information,pimentel2023anticipation}. Chinese reading, with its distinctive writing system and segmentation properties, has also motivated dedicated models of eye-movement control and word processing \citep{rayner2007ezreader,li2020integrated}.

Eye-tracking prediction has expanded from individual high-resource languages to multilingual evaluations \citep{hollenstein2021cmcl,hollenstein2021multilingual,hollenstein2022cmcl,salicchi2022hkamsters}, but regional and under-resourced language varieties remain comparatively underexplored. Cantonese provides an informative case because the value of explicit Cantonese-specific training is itself unsettled in current NLP research. Although Cantonese is widely spoken, its textual resources remain much scarcer than those available for Mandarin, and Cantonese NLP systems frequently rely on transfer from Mandarin-oriented models \citep{jiang2025developing,min2025cantonlu}.

Nevertheless, recent evaluations provide mixed evidence on the need for such adaptation. Chinese-developed LLMs do not necessarily show broader advantages over Western-developed models for languages spoken in China, with their clearest advantage concentrated in Mandarin \citep{wenyi2025chinese}. Cantonese-specific evaluation likewise shows that large general-purpose models can remain highly competitive and that substantial limitations persist in Cantonese linguistic and cultural knowledge \citep{cheng2025hkcanto}. By contrast, Cantonese-adapted models have been found to perform best overall on a recent Cantonese NLU benchmark, although Mandarin models remain competitive on several tasks \citep{min2025cantonlu}; large-scale Cantonese continued pretraining and supervised fine-tuning have also produced strong gains on dedicated Cantonese benchmarks \citep{jiang2025developing}. Taken together, these results suggest that matching model training to Cantonese does not guarantee a uniform advantage: the benefit appears to depend on the adaptation strategy and on the specific evaluation task.

Existing Cantonese NLP studies primarily evaluate performance on linguistic, reasoning, or cultural benchmarks. It remains unclear whether differences between Cantonese-adapted and more general models extend to the probabilistic expectations that best predict human online language processing. We address this question using the Cantonese component of MCFIX \citep{li2023mcfix}, which contains natural reading (NR) and task-specific reading (TSR) measurements for a Cantonese translation of \textit{The Little Prince}. We compare two within-family adaptation contrasts: CKIP GPT-2 Tiny Chinese versus its Cantonese-adapted JED351 derivative, and Qwen2.5-7B versus CantoneseLLM-7B, which was developed from Qwen2.5 through Cantonese-oriented continued pretraining and instruction tuning. For each model, we estimate four information-theoretic predictors: lexical surprisal, POS surprisal, entropy before the target, and entropy reduction. Together, these measures characterize complementary aspects of predictive processing, from uncertainty before the target and lexical or syntactic-category expectation at the target to changes in uncertainty after the target is observed.

Our study addresses three questions. First, do these information-theoretic measures improve prediction beyond conventional lexical, positional, POS, and dependency predictors? Second, within each model family, does Cantonese-specific training improve predictive value, and does the pattern differ between lighter and more extensive adaptation strategies? Third, do these patterns differ across eye-tracking measures, such as first fixation duration (FFD) and total fixation duration (TFD), and between reading tasks?

The main contributions of this study are threefold: (i) a matched comparison of two small and two large language models arranged as two within-family Cantonese-adaptation contrasts; (ii) a joint evaluation of lexical surprisal, POS surprisal, and two entropy-based measures that characterize complementary aspects of predictive processing; and (iii) a five-seed robustness analysis, reading-task comparison, and predictor-block ablations.

\section{Background}

\subsection{Information-theoretic predictors}

We selected four information-theoretic measures to distinguish complementary aspects of model-based expectation around each target word. Entropy before captures uncertainty prior to encountering the target; lexical surprisal quantifies prediction error for the observed word itself; POS surprisal captures prediction error at the syntactic-category level; and entropy reduction measures how observing the target changes uncertainty over the next continuation. We do not assume that these measures map one-to-one onto discrete cognitive stages, but together they characterize pre-target anticipation, target-level expectation, and post-target probabilistic updating.

For a target word $w_t$ and its preceding context $c_t$, \textbf{lexical surprisal} is calculated as:
\begin{equation}
S(w_t)=-\log_2 P(w_t\mid c_t).
\end{equation}
It represents the information conveyed by the observed word: a low-probability continuation has high surprisal. Model-derived lexical surprisal has been widely used to predict reading behavior across languages and eye-tracking settings \citep{hollenstein2021multilingual,salicchi2023surprisal,wilcox2023eleven}.

\textbf{POS surprisal} abstracts away from word identity. Given the target POS $z_t$ and the lexical candidates assigned to that POS, we estimate the aggregate probability of that category and define
\begin{equation}
S(z_t)=-\log P(z_t\mid c_t).
\end{equation}
Unlike a POS dummy, which only records the observed category, POS surprisal quantifies how expected that category is in the current context; related GPT-2-derived POS surprisal has been used to study syntactic-category prediction during natural language comprehension \citep{heilbron2022hierarchy}.

\textbf{Entropy before} measures the dispersion of the model's next-token distribution before the target:
\begin{equation}
H_{\mathrm{before}}=-\sum_{v\in V}P(v\mid c_t)\log_2P(v\mid c_t).
\end{equation}
It captures anticipatory uncertainty rather than the unexpectedness of the word that actually occurs, and contextual entropy has been shown to predict reading-time variation beyond lexical surprisal \citep{pimentel2023anticipation,wilcox2023eleven}.

Finally, we define the next-token \textbf{entropy reduction} metric as
\begin{equation}
ER_t=H_{\mathrm{before}}-H_{\mathrm{after}},
\end{equation}
where $H_{\mathrm{after}}$ is the entropy of the next-token distribution after adding the current word to the context. This operationalization measures the signed change in uncertainty over next-token continuations and should be distinguished from parse-state entropy reduction in probabilistic grammar. Related entropy-reduction measures have been used to model natural reading and naturalistic language comprehension \citep{lowder2018lexical,song2024uncertainty}.

\subsection{Cantonese eye-tracking data}

MCFIX provides parallel Mandarin and Cantonese eye-tracking corpora \citep{li2023mcfix}. The Cantonese corpus was collected from 30 native speakers under two reading conditions: NR, in which participants read for comprehension, and TSR, in which they searched the text for specified information. The dataset includes first fixation duration (FFD), second fixation duration (SFD), and total fixation duration (TFD) at the word level.

The dataset also provides lexical, positional, POS, dependency, and neighbourhood-based annotations used as conventional predictors in our baseline model. These variables are motivated by prior work showing effects of frequency, word length, syntactic structure, and spillover on eye movements during Chinese and multilingual reading \citep{yan2006frequency,zang2018wordlength,vanschijndel2015hierarchic,pollatsek2008immediate}. Cantonese frequency and neighbourhood information derives from Cifu \citep{lai2020cifu}. Li and colleagues further showed that linguistic and GPT-derived predictors are informative for MCFIX.

\section{Data and Models}

\subsection{Data and outcomes}

We used the traditional-character Cantonese portion of MCFIX. The source contained 5,072 NR rows and 5,071 TSR rows. Removing observations without an analyzable lexical target yielded 10,097 rows, matching the reported Cantonese counts of 5,050 (NR) and 5,047 (TSR). To compare all four language models on identical observations, we took the intersection on which every required metric was available. The frozen sample contained 10,011 task-level rows: 5,007 NR and 5,004 TSR observations. The excluded 86 rows account for less than 1\% of the 10,097-row analysis set.

The outcomes were aggregate durations in milliseconds: first fixation duration (FFD), second fixation duration (SFD), and total fixation duration (TFD). NR and TSR observations were analyzed jointly, with reading task included as an indicator predictor. We report both overall cross-validated performance and performance calculated separately on the NR and TSR subsets.

\subsection{Language models}

Table~\ref{tab:models} summarizes the comparison. The model pairs were chosen to provide two Cantonese-adaptation contrasts at different scales, keeping the architectures fixed. Within the small-model pair, JED351 descends from CKIP GPT-2 Tiny Chinese.\footnote{CKIP GPT-2 Tiny Chinese: \url{https://huggingface.co/ckiplab/gpt2-tiny-chinese}} The intermediate JED351 base model\footnote{JED351 base checkpoint: \url{https://huggingface.co/jed351/gpt2-tiny-zh-hk}; Cantonese-Wikipedia checkpoint used in this study: \url{https://huggingface.co/jed351/gpt2_tiny_zh-hk-wiki}} patched the CKIP tokenizer and embedding matrix with Cantonese characters but had not yet been trained on Cantonese; the model used here was subsequently fine-tuned for ten epochs on approximately 50 MB of Cantonese Wikipedia text. Within the large-model pair, CantoneseLLM-7B descends directly from Qwen2.5-7B, allowing a comparison between a broadly pretrained Chinese-capable base model and a derivative receiving substantially more extensive Cantonese-specific training.

\begin{table*}[t]
\centering
\small
\begin{tabular}{@{}llp{10cm}@{}}
\toprule
Model & Model family / scale & Primary characterization \\
\midrule
CKIP & GPT-2 tiny & Traditional-Chinese base model for the small-model contrast \\
JED351 & GPT-2 tiny & CKIP-derived; Cantonese vocabulary patching and Wikipedia fine-tuning \\

Qwen2.5-7B & Qwen2.5, 7B & Multilingual base model for the large-model contrast \\
CantoneseLLM-7B & Qwen2.5, 7B & Qwen2.5-derived; Cantonese continued pretraining and SFT \\
\bottomrule
\end{tabular}
\caption{Autoregressive models evaluated in this study, arranged as two within-family Cantonese-adaptation contrasts. Model descriptions follow the released model repositories and documentation, together with the Qwen2.5 technical report \citep{qwen2024qwen25}.}
\label{tab:models}
\end{table*}

According to the documentation for CantoneseLLM, the model underwent continued pretraining of Qwen2.5-7B on a large corpus drawn from publicly available Hong Kong news and Cantonese websites, followed by supervised instruction tuning on 75,000 pairs \citep{cantoneseLLM}. The comparisons thus represent substantially different Cantonese-adaptation strategies: JED351 combines vocabulary expansion with fine-tuning on a relatively limited amount of Cantonese text, whereas CantoneseLLM-7B combines extensive Cantonese continued pretraining with supervised instruction tuning.

\section{Methods}

\subsection{Metric extraction}

For each target word, the preceding within-sentence context was used to derive four information-theoretic measures from each language model. Multi-token lexical surprisal was calculated as the sum of the conditional surprisals of the target tokens. Entropy before was defined as the entropy of the model's next-token distribution given the preceding context, whereas entropy after was computed after the complete target word was added; entropy reduction was defined as their difference. Lexical surprisal and both entropy measures were expressed in bits.

POS surprisal was constructed using a common candidate inventory from the Cifu Cantonese lexicon \citep{lai2020cifu}. Cifu words were assigned POS categories with PyCantonese \citep{lee2022pycantonese}, and the language-model probabilities of candidates sharing the target POS were aggregated before taking negative log probability. Using the same lexical candidate inventory for all four models makes the POS-surprisal construction directly comparable across models.

\subsection{Regression design}

We used CatBoost regression \citep{prokhorenkova2018catboost}, consistent with the strong performance of gradient-boosting methods with linguistic predictors in prior eye-tracking tasks \citep{hollenstein2021cmcl,salicchi2022hkamsters}. The no-LLM baseline contained reading task, POS dummies, linear distance to root, linear distance to head, dependency depth, neighbourhood size, current and previous frequency, current and previous syllable count, and relative word position. 

For each eye-tracking outcome, we trained: (i) the no-LLM baseline; (ii) four models that added one LLM metric at a time; (iii) a model adding all four metrics; and (iv) leave-one-predictor-block-out models that removed each linguistic or LLM block from the full model. CatBoost used 1,000 iterations and internal random seed 10; all other parameters retained the library defaults.

We used word-level \texttt{KFold} cross-validation with five folds, shuffling, and split seeds 42--46. The same row identities and fold assignments were used for all four language models. For each seed, all reported metrics were calculated from concatenated out-of-fold predictions. Our primary measure was mean absolute error (MAE); we also retained mean squared error, root mean squared error, $R^2$, Pearson correlation, and Spearman correlation.

For a single-metric addition, improvement was defined as
\begin{equation}
\Delta\mathrm{MAE}=\mathrm{MAE}_{\mathrm{baseline}}-\mathrm{MAE}_{\mathrm{augmented}}.
\end{equation}
Positive values indicate lower error. In leave-one-out ablation, contribution was defined as the MAE of the model without a block minus the MAE of the full four-metric model.

As FFD, SFD, and TFD have different baseline scales, we additionally computed a baseline-normalized improvement within each split seed:
\begin{equation}
\%\Delta\mathrm{MAE}=100\left(1-\frac{\mathrm{MAE}_{\mathrm{aug}}}{\mathrm{MAE}_{\mathrm{base}}}\right).
\end{equation}
We summarize this quantity in the appendix. It helps prevent cross-outcome comparisons from being driven solely by the larger numerical scale of the TFD metric.

\section{Results}

\subsection{Incremental value of LLM metrics}

\begin{table*}[t]
\centering
\small
\setlength{\tabcolsep}{4.2pt}
\begin{tabular}{llrrrr}
\toprule
Metric & Measure & JED351 & CKIP & Qwen2.5-7B & CantoneseLLM-7B \\
\midrule
Lexical surprisal & FFD & +0.118 & +0.196 & +0.268 & \textbf{+0.297} \\
 & SFD & +0.064 & +0.081 & +0.152 & \textbf{+0.161} \\
 & TFD & +0.229 & +0.249 & +0.599 & \textbf{+0.683} \\
\addlinespace[1.5pt]
POS surprisal & FFD & +0.118 & +0.200 & +0.276 & \textbf{+0.308} \\
 & SFD & +0.002 & +0.018 & +0.015 & \textbf{+0.050} \\
 & TFD & +0.124 & +0.231 & +0.321 & \textbf{+0.478} \\
\addlinespace[1.5pt]
Entropy before & FFD & +0.147 & +0.112 & +0.163 & \textbf{+0.293} \\
 & SFD & +0.008 & +0.042 & +0.016 & \textbf{+0.097} \\
 & TFD & +0.133 & +0.101 & +0.175 & \textbf{+0.478} \\
\addlinespace[1.5pt]
Entropy reduction & FFD & +0.152 & \textbf{+0.183} & +0.153 & +0.156 \\
 & SFD & -0.008 & \textbf{+0.104} & -0.003 & +0.009 \\
 & TFD & +0.159 & \textbf{+0.362} & +0.163 & +0.162 \\
\addlinespace[1.5pt]
All four & FFD & +0.409 & +0.504 & +0.597 & \textbf{+0.729} \\
 & SFD & +0.125 & +0.197 & +0.200 & \textbf{+0.267} \\
 & TFD & +0.409 & +0.651 & +0.821 & \textbf{+1.251} \\
\bottomrule
\end{tabular}
\caption{Incremental predictive value of each LLM-derived metric across both reading tasks. Values are mean $\Delta$MAE across five split seeds (ms), where $\Delta\mathrm{MAE}=\mathrm{MAE}_{\mathrm{baseline}}-\mathrm{MAE}_{\mathrm{augmented}}$; positive values indicate improvement. The largest gain in each model-comparison row is bold. Full seed-robustness statistics are reported in Appendix Table~\ref{tab:additions-robustness}.}
\label{tab:additions}
\end{table*}

Table~\ref{tab:additions} reports the central comparison. Every model--metric combination improves FFD and TFD on average. Results for SFD are smaller: entropy reduction is approximately null for JED351 and Qwen2.5-7B, but clearly positive for CKIP and slightly positive for CantoneseLLM-7B.

The gains are generally stable across split seeds but modest in absolute predictive terms: even the largest joint improvement is below 2\% of baseline MAE after normalization. They should therefore be interpreted as incremental predictive contributions rather than large changes in overall accuracy.

Lexical surprisal shows a consistent ordering across all three outcomes: CantoneseLLM-7B yields the largest MAE reduction, followed by Qwen2.5-7B, CKIP, and JED351. For TFD, the corresponding gains are 0.683, 0.599, 0.249, and 0.229 ms. POS surprisal and entropy before also generally favor CantoneseLLM-7B. However, the full four-model ordering is clearest for lexical surprisal and does not generalize uniformly across metrics.

Entropy reduction shows a qualitatively different pattern. Although CKIP is not the strongest model for lexical surprisal, its entropy-reduction measure yields the largest mean improvement among all four models for FFD (0.183 ms), SFD (0.104 ms), and TFD (0.362 ms). This dissociation suggests that a model's usefulness for estimating target-word probability need not coincide with its usefulness for estimating how strongly the target changes uncertainty over upcoming material.

\begin{figure*}[t]
\centering
\includegraphics[width=\textwidth]{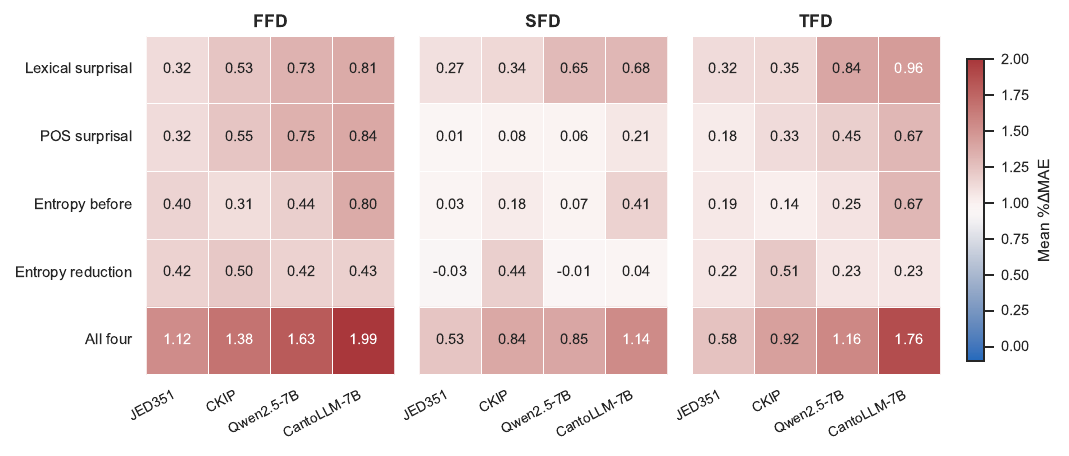}
\caption{Baseline-normalized improvement for each LLM-derived metric and the joint four-metric model. Each cell is mean $\%\Delta$MAE across five split seeds; positive values indicate improvement over the no-LLM baseline. Normalization is performed within each model, outcome, and seed before averaging.}
\label{fig:relative-heatmap}
\end{figure*}

Adding all four measures yields larger improvements than adding any single measure for every model and outcome, consistent with the four metrics providing complementary predictive information. The full CantoneseLLM-7B block reduces MAE by 0.729 ms for FFD, 0.267 ms for SFD, and 1.251 ms for TFD. Within each outcome, the mean joint-model improvements follow the ordering CantoneseLLM-7B $>$ Qwen2.5-7B $>$ CKIP $>$ JED351. Figure~\ref{fig:relative-heatmap} shows the same descriptive ordering after normalization by baseline MAE; exact values are reported in Appendix Table~\ref{tab:relative-additions}. 

Normalization also changes the cross-outcome interpretation: although TFD shows the largest raw reduction for the two 7B models, the relative four-metric improvement is largest for FFD in all four models (1.118--1.991\%). Thus, the apparent magnitude of an outcome-level gain depends on whether it is expressed in milliseconds or relative to baseline error.

The full means, standard deviations, and numbers of positive split-seed effects are reported in Appendix Table~\ref{tab:additions-robustness}.

\subsection{Comparison between reading tasks}

\begin{figure*}[t]
\centering
\includegraphics[width=\textwidth]{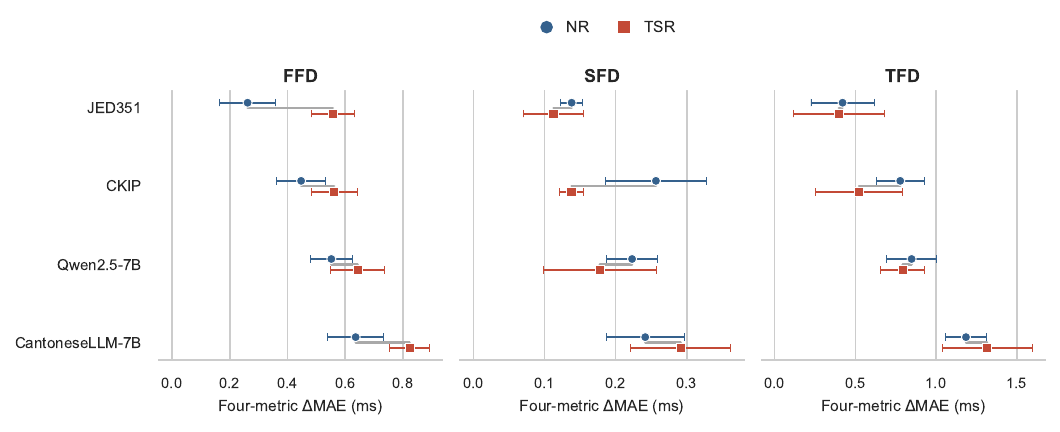}
\caption{Four-metric $\Delta$MAE evaluated separately on NR and TSR observations from the jointly trained models. Points show the five-seed mean and horizontal bars show one standard deviation. }
\label{fig:task-dumbbell}
\end{figure*}

Figure~\ref{fig:task-dumbbell} compares performance on the NR and TSR subsets of the out-of-fold predictions; complete single-metric and joint-model results can be seen in Appendix Tables~\ref{tab:task-nr-full} and~\ref{tab:task-tsr-full}. The task-stratified mean gains show a similar broad ordering: CantoneseLLM-7B produces the largest joint improvement in five of the six task--outcome combinations, and Qwen2.5-7B is generally second. The clearest descriptive task difference concerns FFD. For all four models, the four-metric block yields a larger mean FFD improvement in TSR than in NR; the TSR advantage is 0.296 ms for JED351, 0.114 ms for CKIP, 0.091 ms for Qwen2.5-7B, and 0.187 ms for CantoneseLLM-7B.

The pattern does not generalize uniformly to later measures. CantoneseLLM-7B also yields larger TSR improvements for SFD and TFD, whereas the other three models show slightly or substantially larger NR gains on these outcomes. The contrast is strongest for CKIP, whose joint SFD and TFD improvements are 0.257 and 0.778 ms in NR, compared with 0.138 and 0.523 ms in TSR. Thus, task-stratified differences depend jointly on the eye-tracking measure and the language model: TSR is associated with larger FFD improvements across all four models, whereas SFD and TFD show model-specific patterns.

\subsection{Unique contributions in the full model}

\begin{figure*}[t]
\centering
\includegraphics[width=0.84\textwidth]{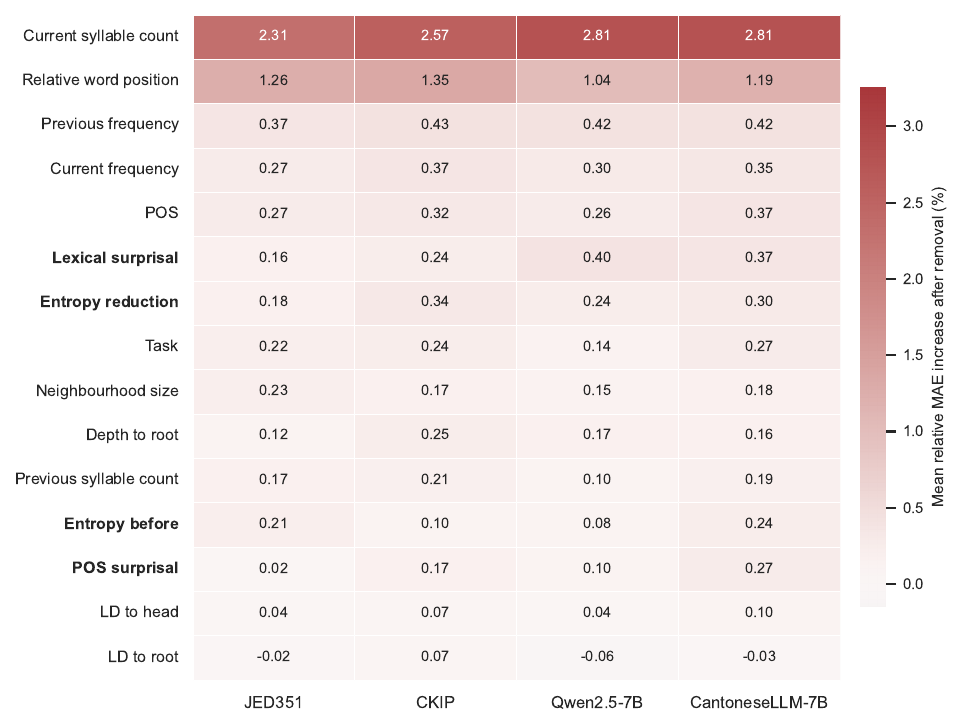}
\caption{Full-model leave-one-predictor-block-out contributions. Within each split seed and eye-tracking outcome, contribution is the MAE increase after removal divided by the complete model's MAE and multiplied by 100. Cells average this percentage across five seeds and the three outcomes; rows are ordered by their mean across models. Positive values indicate that removal worsens prediction, and bold row labels identify LLM-derived metrics.}
\label{fig:full-ablation}
\end{figure*}

Single-metric addition asks whether a metric improves prediction beyond the linguistic baseline, whereas ablation asks whether that metric retains a unique contribution when the other three LLM-derived measures are already present. Figure~\ref{fig:full-ablation} extends the ablation analysis to all linguistic and LLM predictor blocks. Current syllable count and relative word position rank first and second for all four models. Among the LLM-derived blocks, lexical surprisal has the largest mean conditional contribution for Qwen2.5-7B and CantoneseLLM-7B, entropy reduction for CKIP, and entropy before for JED351. Overall, the LLM-derived variables occupy the middle of the ranking, complementing rather than replacing the strongest form- and position-based predictors. Outcome-specific raw LLM ablations are reported in Appendix Table~\ref{tab:llm-ablation}.

The conditional contributions of the LLM-derived measures are generally smaller and less uniform than their single-metric addition gains, consistent with partial overlap among the four measures. Appendix Figure~\ref{fig:metric-correlations} quantifies this overlap. The strongest and most consistent association is, unsurprisingly, between entropy before and entropy reduction ($\rho=.55$--$.67$ across models), whereas most other pairwise correlations are weak. CantoneseLLM additionally shows moderate associations of lexical surprisal with entropy before ($\rho=.56$) and entropy reduction ($\rho=.38$). 
The four measures therefore share information without being interchangeable.

\section{Discussion}

\subsection{Do LLM-derived metrics improve prediction beyond conventional predictors?}

Our first question asked whether the four information-theoretic measures provide predictive information beyond conventional lexical, positional, POS, and dependency predictors. The results indicate that they do, although the gains are modest. Each metric improves prediction in at least some model--outcome combinations, and the joint four-metric model outperforms every corresponding single-metric model. At the same time, even the largest baseline-normalized joint improvement remains below 2\%, indicating that these measures provide incremental rather than transformative gains in predictive accuracy.

The ablation analysis further clarifies their role. Current syllable count and relative word position remain the strongest predictor blocks across models, while the LLM-derived measures generally occupy the middle of the ranking. The information-theoretic measures therefore complement rather than replace established linguistic predictors. Their conditional contributions are also smaller than their single-metric additions, consistent with partial overlap among the four measures.

These findings support the use of multiple theoretically distinct probability-derived measures when examining the relationship between language-model expectations and naturalistic reading behavior.

\subsection{How does predictive value vary with Cantonese-specific training?}

Our second question concerned whether Cantonese-specific training improves predictive value within each model family, and whether the pattern differs between lighter and more extensive adaptation strategies. The results support a qualified association rather than a uniform advantage from Cantonese adaptation. In the small-model comparison, JED351 does not show a general predictive advantage over its CKIP base model despite Cantonese vocabulary expansion and Wikipedia fine-tuning. In contrast, CantoneseLLM-7B, which underwent substantially more extensive Cantonese continued pretraining and supervised instruction tuning, yields the largest mean lexical-surprisal and joint-model improvements across all three eye-tracking outcomes.

For the joint four-metric model, the JED351--CKIP contrast is negative for FFD, SFD, and TFD, providing no evidence of an overall advantage from the lighter Cantonese adaptation strategy. The large-model pair shows the opposite pattern: CantoneseLLM-7B exceeds Qwen2.5-7B by $0.132$, $0.066$, and $0.430$ ms for FFD, SFD, and TFD, respectively, with positive contrasts across all five split seeds. Entropy before shows similarly consistent advantages, whereas the entropy-reduction contrasts remain close to zero. Lexical and POS surprisal also favor CantoneseLLM-7B on average, although less uniformly across seeds. The complete within-family contrasts are shown in Appendix Figure~\ref{fig:within-family-adaptation}.

These comparisons do not isolate the effect of Cantonese training from every other training difference between the checkpoints. Within each pair, Cantonese adaptation involves multiple changes, including vocabulary modification, continued pretraining, or instruction tuning. Nor should the stronger performance of the 7B models be interpreted as evidence that scale alone improves psycholinguistic fit: more training and larger models have been reported to worsen surprisal fit in controlled English studies \citep{oh2023twobillion,oh2024frequency}, and similar results were found for other languages (e.g. Brazilian Portuguese, \citet{alves2025benchmarking}). The present results therefore suggest that more extensive Cantonese-specific training can be associated with stronger predictive fit, while Cantonese adaptation by itself is not sufficient to guarantee an advantage.

\subsection{Do the effects vary across eye-tracking measures and reading tasks?}

Our third question asked whether the predictive patterns vary across eye-tracking outcomes and between natural and task-specific reading. The strongest raw improvements often occur for TFD, but this partly reflects its larger baseline-error scale: after baseline normalization, the joint four-metric improvement is largest for FFD in every model. Because FFD indexes the first fixation whereas SFD records a second fixation, these outcomes plausibly emphasize different stages of processing; second-pass fixation measures can be more sensitive to later structural demands \citep{conklin2016using,li2023mcfix}.

The clearest task-related pattern concerns FFD. All four models show larger mean joint-model improvements in TSR than in NR for this measure, which is compatible with goal-directed information search increasing the relevance of lexical expectations during initial processing. However, task-specific conditioning does not necessarily improve fit \citep{grutekeklein2024information}, and the pattern does not extend uniformly to SFD or TFD. CantoneseLLM-7B shows larger TSR gains for these later measures, whereas the other models often show larger NR gains.

The results suggest that the predictive value of LLM-derived information-theoretic measures may depend on both the reading objective and the eye-tracking measure being modeled.

\section{Conclusion}

This study examined whether Cantonese-specific language-model training is associated with stronger prediction of Cantonese reading behavior. Across four information-theoretic measures, LLM-derived predictors provided consistent but modest improvements beyond conventional linguistic predictors. More importantly, Cantonese adaptation did not yield a uniform advantage: the lightly adapted JED351 model did not outperform its CKIP base model overall, whereas the more extensively adapted CantoneseLLM-7B consistently outperformed Qwen2.5-7B for lexical surprisal and the joint four-metric model. At the same time, entropy reduction showed a different cross-model pattern, with CKIP yielding the strongest gains. 

Together, these results suggest that psycholinguistic alignment depends not simply on whether a model has been trained on more Cantonese, but on both the adaptation strategy and the aspect of predictive uncertainty being modeled. Better prediction of the upcoming word therefore does not necessarily imply better modeling of how linguistic input updates uncertainty during reading.

\section*{Limitations}

The model comparisons do not isolate the effect of Cantonese training. JED351 differs from CKIP in vocabulary, embeddings, and Cantonese Wikipedia fine-tuning, while CantoneseLLM-7B differs from Qwen2.5-7B in continued Cantonese pretraining and instruction tuning; the two pairs also differ in architecture and scale. The results therefore reflect broader adaptation strategies rather than any single training component.

The extraction pipeline tokenizes context and target separately, imposing a lexical boundary that differs from natural concatenated tokenization on 246 of 5,072 Qwen-family rows. The data are further limited to one literary translation and participant sample, and entropy values are not directly comparable across model--tokenizer systems with different continuation spaces.

\bibliography{references}

\appendix

\section{Reproducibility Details}
\label{sec:reproducibility}

The regression analysis uses split seeds 42--46, five shuffled folds per seed, CatBoost internal seed 10, and 1,000 boosting iterations. Code will be released upon acceptance.

\FloatBarrier

\section{Supplementary Analyses}
\label{sec:supplementary}


\begin{table*}[t]
\centering
\small
\setlength{\tabcolsep}{2.7pt}
\renewcommand{\arraystretch}{1.02}
\resizebox{0.80\textwidth}{!}{%
\begin{tabular}{llrrrr}
\toprule
Metric & Measure & JED351 & CKIP & Qwen2.5-7B & CantoneseLLM-7B \\
\midrule

Lexical surprisal
& FFD
& \shortstack{+0.118 $\pm$ 0.031\\(5/5)}
& \shortstack{+0.196 $\pm$ 0.048\\(5/5)}
& \shortstack{+0.268 $\pm$ 0.079\\(5/5)}
& \shortstack{+0.297 $\pm$ 0.054\\(5/5)} \\

& SFD
& \shortstack{+0.064 $\pm$ 0.016\\(5/5)}
& \shortstack{+0.081 $\pm$ 0.020\\(5/5)}
& \shortstack{+0.152 $\pm$ 0.030\\(5/5)}
& \shortstack{+0.161 $\pm$ 0.025\\(5/5)} \\

& TFD
& \shortstack{+0.229 $\pm$ 0.036\\(5/5)}
& \shortstack{+0.249 $\pm$ 0.112\\(5/5)}
& \shortstack{+0.599 $\pm$ 0.043\\(5/5)}
& \shortstack{+0.683 $\pm$ 0.066\\(5/5)} \\

\addlinespace[1.0pt]

POS surprisal
& FFD
& \shortstack{+0.118 $\pm$ 0.057\\(5/5)}
& \shortstack{+0.200 $\pm$ 0.035\\(5/5)}
& \shortstack{+0.276 $\pm$ 0.032\\(5/5)}
& \shortstack{+0.308 $\pm$ 0.037\\(5/5)} \\

& SFD
& \shortstack{+0.002 $\pm$ 0.026\\(3/5)}
& \shortstack{+0.018 $\pm$ 0.010\\(5/5)}
& \shortstack{+0.015 $\pm$ 0.037\\(3/5)}
& \shortstack{+0.050 $\pm$ 0.020\\(5/5)} \\

& TFD
& \shortstack{+0.124 $\pm$ 0.111\\(4/5)}
& \shortstack{+0.231 $\pm$ 0.015\\(5/5)}
& \shortstack{+0.321 $\pm$ 0.076\\(5/5)}
& \shortstack{+0.478 $\pm$ 0.054\\(5/5)} \\

\addlinespace[1.0pt]

Entropy before
& FFD
& \shortstack{+0.147 $\pm$ 0.039\\(5/5)}
& \shortstack{+0.112 $\pm$ 0.040\\(5/5)}
& \shortstack{+0.163 $\pm$ 0.054\\(5/5)}
& \shortstack{+0.293 $\pm$ 0.023\\(5/5)} \\

& SFD
& \shortstack{+0.008 $\pm$ 0.013\\(3/5)}
& \shortstack{+0.042 $\pm$ 0.042\\(4/5)}
& \shortstack{+0.016 $\pm$ 0.017\\(5/5)}
& \shortstack{+0.097 $\pm$ 0.012\\(5/5)} \\

& TFD
& \shortstack{+0.133 $\pm$ 0.153\\(4/5)}
& \shortstack{+0.101 $\pm$ 0.082\\(4/5)}
& \shortstack{+0.175 $\pm$ 0.090\\(5/5)}
& \shortstack{+0.478 $\pm$ 0.060\\(5/5)} \\

\addlinespace[1.0pt]

Entropy reduction
& FFD
& \shortstack{+0.152 $\pm$ 0.013\\(5/5)}
& \shortstack{+0.183 $\pm$ 0.038\\(5/5)}
& \shortstack{+0.153 $\pm$ 0.027\\(5/5)}
& \shortstack{+0.156 $\pm$ 0.027\\(5/5)} \\

& SFD
& \shortstack{-0.008 $\pm$ 0.016\\(2/5)}
& \shortstack{+0.104 $\pm$ 0.032\\(5/5)}
& \shortstack{-0.003 $\pm$ 0.011\\(2/5)}
& \shortstack{+0.009 $\pm$ 0.032\\(3/5)} \\

& TFD
& \shortstack{+0.159 $\pm$ 0.085\\(5/5)}
& \shortstack{+0.362 $\pm$ 0.090\\(5/5)}
& \shortstack{+0.163 $\pm$ 0.064\\(5/5)}
& \shortstack{+0.162 $\pm$ 0.085\\(5/5)} \\

\addlinespace[1.0pt]

All four
& FFD
& \shortstack{+0.409 $\pm$ 0.078\\(5/5)}
& \shortstack{+0.504 $\pm$ 0.067\\(5/5)}
& \shortstack{+0.597 $\pm$ 0.078\\(5/5)}
& \shortstack{+0.729 $\pm$ 0.074\\(5/5)} \\

& SFD
& \shortstack{+0.125 $\pm$ 0.025\\(5/5)}
& \shortstack{+0.197 $\pm$ 0.039\\(5/5)}
& \shortstack{+0.200 $\pm$ 0.030\\(5/5)}
& \shortstack{+0.267 $\pm$ 0.034\\(5/5)} \\

& TFD
& \shortstack{+0.409 $\pm$ 0.201\\(5/5)}
& \shortstack{+0.651 $\pm$ 0.193\\(5/5)}
& \shortstack{+0.821 $\pm$ 0.136\\(5/5)}
& \shortstack{+1.251 $\pm$ 0.173\\(5/5)} \\

\bottomrule
\end{tabular}
}

\caption{Five-seed robustness of the pooled single-metric and joint-model additions. Each cell reports mean $\Delta\mathrm{MAE} \pm \mathrm{SD}$ in milliseconds; the second line gives the number of split seeds with $\Delta\mathrm{MAE}>0$ out of five. The count is a descriptive stability summary, not a significance test.}
\label{tab:additions-robustness}
\end{table*}


\begin{table*}[t]
\centering
\small
\setlength{\tabcolsep}{2.8pt}

\begin{tabular}{llrrrr}
\toprule
Metric & Measure & JED351 & CKIP & Qwen2.5-7B & CantoneseLLM-7B \\
\midrule

Lexical surprisal
& FFD & +0.323 $\pm$ 0.084 & +0.534 $\pm$ 0.131 & +0.732 $\pm$ 0.215 & +0.810 $\pm$ 0.147 \\
& SFD & +0.271 $\pm$ 0.069 & +0.345 $\pm$ 0.084 & +0.649 $\pm$ 0.128 & +0.684 $\pm$ 0.105 \\
& TFD & +0.323 $\pm$ 0.049 & +0.350 $\pm$ 0.158 & +0.844 $\pm$ 0.063 & +0.963 $\pm$ 0.093 \\

\addlinespace[1.2pt]

POS surprisal
& FFD & +0.323 $\pm$ 0.154 & +0.547 $\pm$ 0.096 & +0.752 $\pm$ 0.086 & +0.842 $\pm$ 0.098 \\
& SFD & +0.007 $\pm$ 0.111 & +0.076 $\pm$ 0.041 & +0.063 $\pm$ 0.157 & +0.213 $\pm$ 0.085 \\
& TFD & +0.175 $\pm$ 0.156 & +0.325 $\pm$ 0.022 & +0.453 $\pm$ 0.107 & +0.675 $\pm$ 0.076 \\

\addlinespace[1.2pt]

Entropy before
& FFD & +0.401 $\pm$ 0.105 & +0.306 $\pm$ 0.109 & +0.445 $\pm$ 0.146 & +0.801 $\pm$ 0.063 \\
& SFD & +0.033 $\pm$ 0.053 & +0.179 $\pm$ 0.177 & +0.069 $\pm$ 0.072 & +0.411 $\pm$ 0.051 \\
& TFD & +0.187 $\pm$ 0.215 & +0.142 $\pm$ 0.116 & +0.247 $\pm$ 0.126 & +0.674 $\pm$ 0.085 \\

\addlinespace[1.2pt]

Entropy reduction
& FFD & +0.416 $\pm$ 0.035 & +0.500 $\pm$ 0.103 & +0.417 $\pm$ 0.075 & +0.426 $\pm$ 0.073 \\
& SFD & -0.032 $\pm$ 0.067 & +0.443 $\pm$ 0.136 & -0.012 $\pm$ 0.045 & +0.039 $\pm$ 0.137 \\
& TFD & +0.225 $\pm$ 0.120 & +0.511 $\pm$ 0.126 & +0.230 $\pm$ 0.090 & +0.229 $\pm$ 0.119 \\

\addlinespace[1.2pt]

All four
& FFD & +1.118 $\pm$ 0.212 & +1.377 $\pm$ 0.183 & +1.631 $\pm$ 0.208 & +1.991 $\pm$ 0.200 \\
& SFD & +0.534 $\pm$ 0.104 & +0.839 $\pm$ 0.164 & +0.853 $\pm$ 0.127 & +1.135 $\pm$ 0.147 \\
& TFD & +0.577 $\pm$ 0.282 & +0.917 $\pm$ 0.272 & +1.157 $\pm$ 0.190 & +1.764 $\pm$ 0.244 \\

\bottomrule
\end{tabular}

\caption{Baseline-normalized improvement for each LLM-derived metric. Values are mean percentage $\Delta$MAE $\pm$ SD across the five split seeds, calculated within seed as $100(\mathrm{MAE}_{\mathrm{baseline}}-\mathrm{MAE}_{\mathrm{augmented}})/\mathrm{MAE}_{\mathrm{baseline}}$. Positive values indicate improvement.}
\label{tab:relative-additions}
\end{table*}


\begin{table*}[t]
\centering
\small
\setlength{\tabcolsep}{2.7pt}
\renewcommand{\arraystretch}{1.02}
\resizebox{\textwidth}{!}{%
\begin{tabular}{llrrrr}
\toprule
Metric & Measure & JED351 & CKIP & Qwen2.5-7B & CantoneseLLM-7B \\
\midrule

Lexical surprisal
& FFD
& \shortstack{+0.045 $\pm$ 0.072\\(4/5)}
& \shortstack{+0.204 $\pm$ 0.093\\(5/5)}
& \shortstack{+0.239 $\pm$ 0.094\\(5/5)}
& \shortstack{+0.237 $\pm$ 0.069\\(5/5)} \\

& SFD
& \shortstack{+0.063 $\pm$ 0.012\\(5/5)}
& \shortstack{+0.105 $\pm$ 0.034\\(5/5)}
& \shortstack{+0.147 $\pm$ 0.025\\(5/5)}
& \shortstack{+0.131 $\pm$ 0.024\\(5/5)} \\

& TFD
& \shortstack{+0.115 $\pm$ 0.152\\(3/5)}
& \shortstack{+0.261 $\pm$ 0.181\\(5/5)}
& \shortstack{+0.555 $\pm$ 0.110\\(5/5)}
& \shortstack{+0.625 $\pm$ 0.117\\(5/5)} \\

\addlinespace[1.0pt]

POS surprisal
& FFD
& \shortstack{+0.066 $\pm$ 0.082\\(4/5)}
& \shortstack{+0.185 $\pm$ 0.069\\(5/5)}
& \shortstack{+0.274 $\pm$ 0.025\\(5/5)}
& \shortstack{+0.309 $\pm$ 0.057\\(5/5)} \\

& SFD
& \shortstack{+0.012 $\pm$ 0.031\\(3/5)}
& \shortstack{+0.016 $\pm$ 0.044\\(4/5)}
& \shortstack{+0.030 $\pm$ 0.042\\(4/5)}
& \shortstack{+0.061 $\pm$ 0.024\\(5/5)} \\

& TFD
& \shortstack{+0.147 $\pm$ 0.175\\(4/5)}
& \shortstack{+0.213 $\pm$ 0.059\\(5/5)}
& \shortstack{+0.299 $\pm$ 0.147\\(5/5)}
& \shortstack{+0.473 $\pm$ 0.048\\(5/5)} \\

\addlinespace[1.0pt]

Entropy before
& FFD
& \shortstack{+0.140 $\pm$ 0.076\\(5/5)}
& \shortstack{+0.083 $\pm$ 0.030\\(5/5)}
& \shortstack{+0.139 $\pm$ 0.097\\(5/5)}
& \shortstack{+0.248 $\pm$ 0.046\\(5/5)} \\

& SFD
& \shortstack{-0.001 $\pm$ 0.029\\(2/5)}
& \shortstack{+0.058 $\pm$ 0.059\\(4/5)}
& \shortstack{+0.035 $\pm$ 0.032\\(5/5)}
& \shortstack{+0.089 $\pm$ 0.031\\(5/5)} \\

& TFD
& \shortstack{+0.138 $\pm$ 0.089\\(5/5)}
& \shortstack{+0.133 $\pm$ 0.152\\(4/5)}
& \shortstack{+0.188 $\pm$ 0.132\\(5/5)}
& \shortstack{+0.452 $\pm$ 0.111\\(5/5)} \\

\addlinespace[1.0pt]

Entropy reduction
& FFD
& \shortstack{+0.180 $\pm$ 0.026\\(5/5)}
& \shortstack{+0.201 $\pm$ 0.029\\(5/5)}
& \shortstack{+0.177 $\pm$ 0.051\\(5/5)}
& \shortstack{+0.152 $\pm$ 0.038\\(5/5)} \\

& SFD
& \shortstack{+0.032 $\pm$ 0.020\\(5/5)}
& \shortstack{+0.127 $\pm$ 0.057\\(5/5)}
& \shortstack{+0.028 $\pm$ 0.019\\(5/5)}
& \shortstack{+0.027 $\pm$ 0.039\\(4/5)} \\

& TFD
& \shortstack{+0.394 $\pm$ 0.189\\(5/5)}
& \shortstack{+0.443 $\pm$ 0.081\\(5/5)}
& \shortstack{+0.254 $\pm$ 0.164\\(5/5)}
& \shortstack{+0.243 $\pm$ 0.123\\(5/5)} \\

\addlinespace[1.0pt]

All four
& FFD
& \shortstack{+0.262 $\pm$ 0.096\\(5/5)}
& \shortstack{+0.447 $\pm$ 0.084\\(5/5)}
& \shortstack{+0.552 $\pm$ 0.073\\(5/5)}
& \shortstack{+0.636 $\pm$ 0.097\\(5/5)} \\

& SFD
& \shortstack{+0.138 $\pm$ 0.015\\(5/5)}
& \shortstack{+0.257 $\pm$ 0.071\\(5/5)}
& \shortstack{+0.223 $\pm$ 0.035\\(5/5)}
& \shortstack{+0.242 $\pm$ 0.055\\(5/5)} \\

& TFD
& \shortstack{+0.421 $\pm$ 0.196\\(5/5)}
& \shortstack{+0.778 $\pm$ 0.150\\(5/5)}
& \shortstack{+0.849 $\pm$ 0.156\\(5/5)}
& \shortstack{+1.186 $\pm$ 0.127\\(5/5)} \\

\bottomrule
\end{tabular}
}

\caption{Complete NR task-stratified additions from the pooled models. Each cell reports mean $\Delta$MAE $\pm$ SD in milliseconds, followed by the number of positive split-seed effects out of five. These are evaluations of NR out-of-fold observations from jointly trained NR+TSR models, not separately trained NR-only models.}
\label{tab:task-nr-full}
\end{table*}


\begin{figure*}[p]
\centering

\begin{minipage}[t]{0.49\textwidth}
\centering
\includegraphics[width=\linewidth]
  {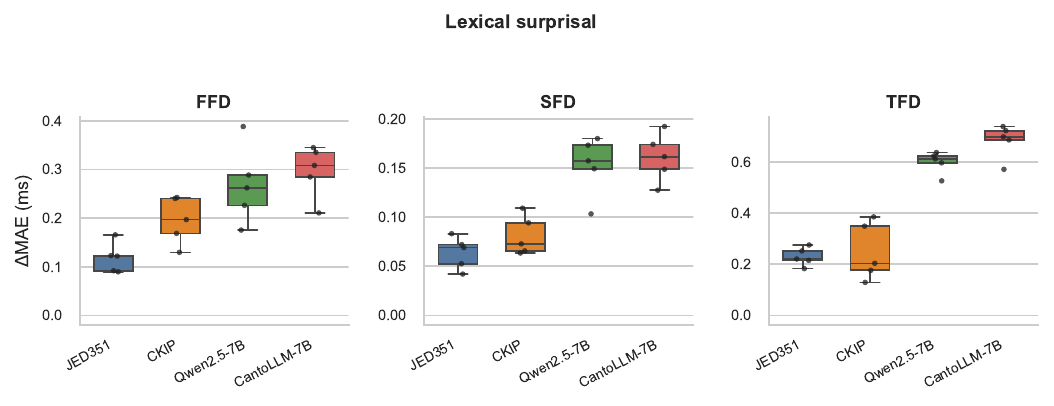}
\textbf{(a) Lexical surprisal}
\end{minipage}
\hfill
\begin{minipage}[t]{0.49\textwidth}
\centering
\includegraphics[width=\linewidth]
  {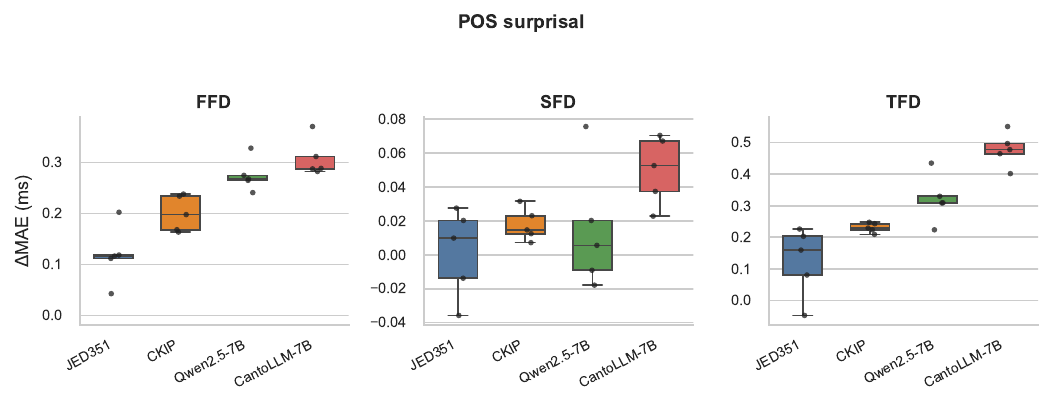}
\textbf{(b) POS surprisal}
\end{minipage}

\vspace{0.6em}

\begin{minipage}[t]{0.49\textwidth}
\centering
\includegraphics[width=\linewidth]
  {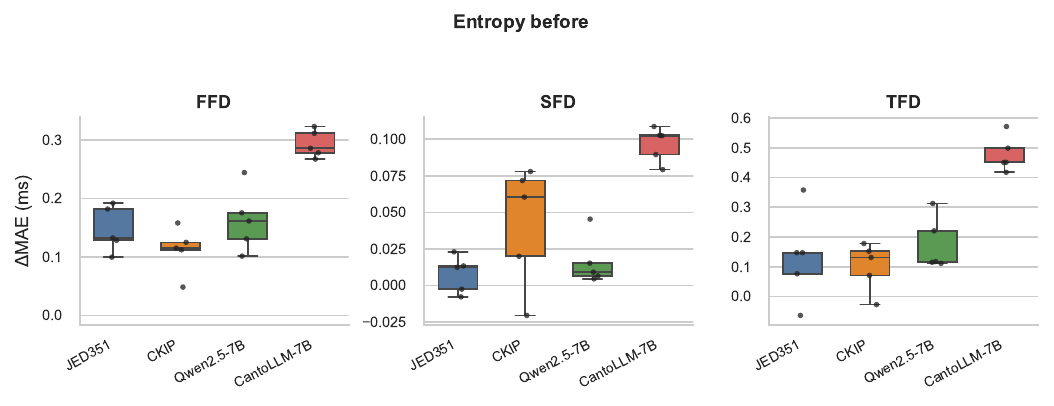}
\textbf{(c) Entropy before}
\end{minipage}
\hfill
\begin{minipage}[t]{0.49\textwidth}
\centering
\includegraphics[width=\linewidth]
  {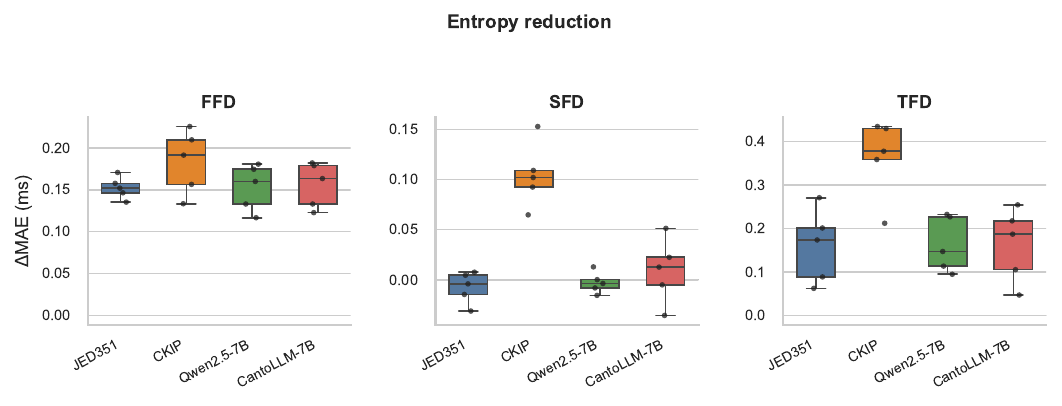}
\textbf{(d) Entropy reduction}
\end{minipage}

\caption{Seed-level $\Delta$MAE distributions for the four LLM-derived metrics. Each panel compares JED351, CKIP, Qwen2.5-7B, and CantoneseLLM-7B. Each box contains the five split-seed values, which are also shown as individual points. Positive values indicate improvement over the no-LLM baseline.}
\label{fig:seed-metric-boxplots}
\end{figure*}

\begin{figure*}[t]
\centering
\includegraphics[width=\textwidth]{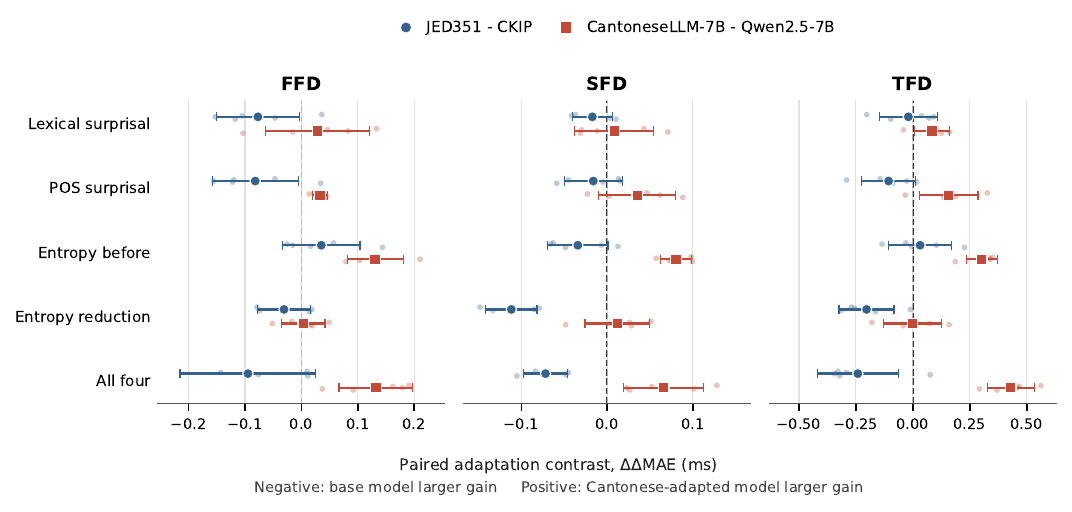}
\caption{Within-family Cantonese-adaptation contrasts. Values show the paired difference in $\Delta$MAE between each Cantonese-adapted model and its base counterpart across the five split seeds. Negative values indicate larger predictive gains for the base model, whereas positive values indicate larger gains for the Cantonese-adapted model.}
\label{fig:within-family-adaptation}
\end{figure*}


\begin{table*}[t]
\centering
\small
\setlength{\tabcolsep}{2.7pt}
\renewcommand{\arraystretch}{1.02}
\resizebox{\textwidth}{!}{%
\begin{tabular}{llrrrr}
\toprule
Metric & Measure & JED351 & CKIP & Qwen2.5-7B & CantoneseLLM-7B \\
\midrule

Lexical surprisal
& FFD
& \shortstack{+0.192 $\pm$ 0.037\\(5/5)}
& \shortstack{+0.188 $\pm$ 0.018\\(5/5)}
& \shortstack{+0.298 $\pm$ 0.071\\(5/5)}
& \shortstack{+0.356 $\pm$ 0.067\\(5/5)} \\

& SFD
& \shortstack{+0.064 $\pm$ 0.025\\(5/5)}
& \shortstack{+0.057 $\pm$ 0.030\\(5/5)}
& \shortstack{+0.157 $\pm$ 0.040\\(5/5)}
& \shortstack{+0.190 $\pm$ 0.057\\(5/5)} \\

& TFD
& \shortstack{+0.344 $\pm$ 0.112\\(5/5)}
& \shortstack{+0.237 $\pm$ 0.210\\(4/5)}
& \shortstack{+0.642 $\pm$ 0.085\\(5/5)}
& \shortstack{+0.741 $\pm$ 0.098\\(5/5)} \\

\addlinespace[1.0pt]

POS surprisal
& FFD
& \shortstack{+0.171 $\pm$ 0.035\\(5/5)}
& \shortstack{+0.216 $\pm$ 0.043\\(5/5)}
& \shortstack{+0.277 $\pm$ 0.045\\(5/5)}
& \shortstack{+0.308 $\pm$ 0.038\\(5/5)} \\

& SFD
& \shortstack{-0.009 $\pm$ 0.044\\(2/5)}
& \shortstack{+0.020 $\pm$ 0.047\\(3/5)}
& \shortstack{+0.000 $\pm$ 0.039\\(2/5)}
& \shortstack{+0.039 $\pm$ 0.034\\(5/5)} \\

& TFD
& \shortstack{+0.101 $\pm$ 0.146\\(3/5)}
& \shortstack{+0.248 $\pm$ 0.081\\(5/5)}
& \shortstack{+0.344 $\pm$ 0.060\\(5/5)}
& \shortstack{+0.484 $\pm$ 0.105\\(5/5)} \\

\addlinespace[1.0pt]

Entropy before
& FFD
& \shortstack{+0.154 $\pm$ 0.045\\(5/5)}
& \shortstack{+0.140 $\pm$ 0.063\\(5/5)}
& \shortstack{+0.186 $\pm$ 0.040\\(5/5)}
& \shortstack{+0.339 $\pm$ 0.049\\(5/5)} \\

& SFD
& \shortstack{+0.017 $\pm$ 0.031\\(4/5)}
& \shortstack{+0.026 $\pm$ 0.034\\(4/5)}
& \shortstack{-0.003 $\pm$ 0.045\\(4/5)}
& \shortstack{+0.105 $\pm$ 0.028\\(5/5)} \\

& TFD
& \shortstack{+0.127 $\pm$ 0.229\\(4/5)}
& \shortstack{+0.068 $\pm$ 0.161\\(3/5)}
& \shortstack{+0.162 $\pm$ 0.115\\(5/5)}
& \shortstack{+0.504 $\pm$ 0.137\\(5/5)} \\

\addlinespace[1.0pt]

Entropy reduction
& FFD
& \shortstack{+0.125 $\pm$ 0.011\\(5/5)}
& \shortstack{+0.165 $\pm$ 0.066\\(5/5)}
& \shortstack{+0.129 $\pm$ 0.043\\(5/5)}
& \shortstack{+0.159 $\pm$ 0.030\\(5/5)} \\

& SFD
& \shortstack{-0.047 $\pm$ 0.034\\(1/5)}
& \shortstack{+0.081 $\pm$ 0.029\\(5/5)}
& \shortstack{-0.033 $\pm$ 0.029\\(0/5)}
& \shortstack{-0.008 $\pm$ 0.045\\(2/5)} \\

& TFD
& \shortstack{-0.076 $\pm$ 0.117\\(2/5)}
& \shortstack{+0.281 $\pm$ 0.218\\(4/5)}
& \shortstack{+0.072 $\pm$ 0.128\\(3/5)}
& \shortstack{+0.081 $\pm$ 0.081\\(4/5)} \\

\addlinespace[1.0pt]

All four
& FFD
& \shortstack{+0.557 $\pm$ 0.074\\(5/5)}
& \shortstack{+0.562 $\pm$ 0.080\\(5/5)}
& \shortstack{+0.643 $\pm$ 0.093\\(5/5)}
& \shortstack{+0.823 $\pm$ 0.069\\(5/5)} \\

& SFD
& \shortstack{+0.113 $\pm$ 0.042\\(5/5)}
& \shortstack{+0.138 $\pm$ 0.017\\(5/5)}
& \shortstack{+0.178 $\pm$ 0.080\\(5/5)}
& \shortstack{+0.292 $\pm$ 0.070\\(5/5)} \\

& TFD
& \shortstack{+0.397 $\pm$ 0.280\\(5/5)}
& \shortstack{+0.523 $\pm$ 0.270\\(5/5)}
& \shortstack{+0.793 $\pm$ 0.136\\(5/5)}
& \shortstack{+1.317 $\pm$ 0.279\\(5/5)} \\

\bottomrule
\end{tabular}
}

\caption{Complete TSR task-stratified additions from the pooled models. Each cell reports mean $\Delta$MAE $\pm$ SD in milliseconds, followed by the number of positive split-seed effects out of five. These are evaluations of TSR out-of-fold observations from jointly trained NR+TSR models, not separately trained TSR-only models.}
\label{tab:task-tsr-full}
\end{table*}


\begin{table*}[t]
\centering
\small
\setlength{\tabcolsep}{4.2pt}

\begin{tabular}{llrrrr}
\toprule
Removed block & Measure & JED351 & CKIP & Qwen2.5-7B & CantoneseLLM-7B \\
\midrule

Lexical surprisal
& FFD & +0.050 $\pm$ 0.029 & +0.083 $\pm$ 0.026 & +0.088 $\pm$ 0.017 & +0.083 $\pm$ 0.072 \\
& SFD & +0.073 $\pm$ 0.020 & +0.061 $\pm$ 0.024 & +0.120 $\pm$ 0.036 & +0.115 $\pm$ 0.019 \\
& TFD & +0.020 $\pm$ 0.190 & +0.166 $\pm$ 0.134 & +0.313 $\pm$ 0.083 & +0.269 $\pm$ 0.147 \\

\addlinespace[1.5pt]

POS surprisal
& FFD & +0.027 $\pm$ 0.064 & +0.084 $\pm$ 0.052 & +0.097 $\pm$ 0.053 & +0.146 $\pm$ 0.059 \\
& SFD & +0.001 $\pm$ 0.027 & +0.026 $\pm$ 0.028 & -0.002 $\pm$ 0.032 & +0.029 $\pm$ 0.028 \\
& TFD & -0.003 $\pm$ 0.134 & +0.112 $\pm$ 0.157 & +0.038 $\pm$ 0.141 & +0.188 $\pm$ 0.060 \\

\addlinespace[1.5pt]

Entropy before
& FFD & +0.091 $\pm$ 0.053 & +0.070 $\pm$ 0.046 & +0.063 $\pm$ 0.065 & +0.072 $\pm$ 0.043 \\
& SFD & +0.050 $\pm$ 0.015 & +0.029 $\pm$ 0.023 & +0.023 $\pm$ 0.025 & +0.052 $\pm$ 0.026 \\
& TFD & +0.123 $\pm$ 0.122 & -0.021 $\pm$ 0.137 & -0.029 $\pm$ 0.118 & +0.215 $\pm$ 0.044 \\

\addlinespace[1.5pt]

Entropy reduction
& FFD & +0.102 $\pm$ 0.030 & +0.148 $\pm$ 0.043 & +0.143 $\pm$ 0.044 & +0.163 $\pm$ 0.088 \\
& SFD & +0.016 $\pm$ 0.018 & +0.079 $\pm$ 0.018 & +0.047 $\pm$ 0.033 & +0.036 $\pm$ 0.043 \\
& TFD & +0.132 $\pm$ 0.099 & +0.185 $\pm$ 0.096 & +0.086 $\pm$ 0.195 & +0.213 $\pm$ 0.107 \\

\bottomrule
\end{tabular}

\caption{Leave-one-LLM-metric-out ablation from each full four-metric model. Values are mean increases in MAE after removal $\pm$ SD across split seeds (ms); positive values indicate a unique contribution conditional on all other predictors.}
\label{tab:llm-ablation}
\end{table*}


\begin{figure*}[t]
\centering
\includegraphics[width=0.88\textwidth]
{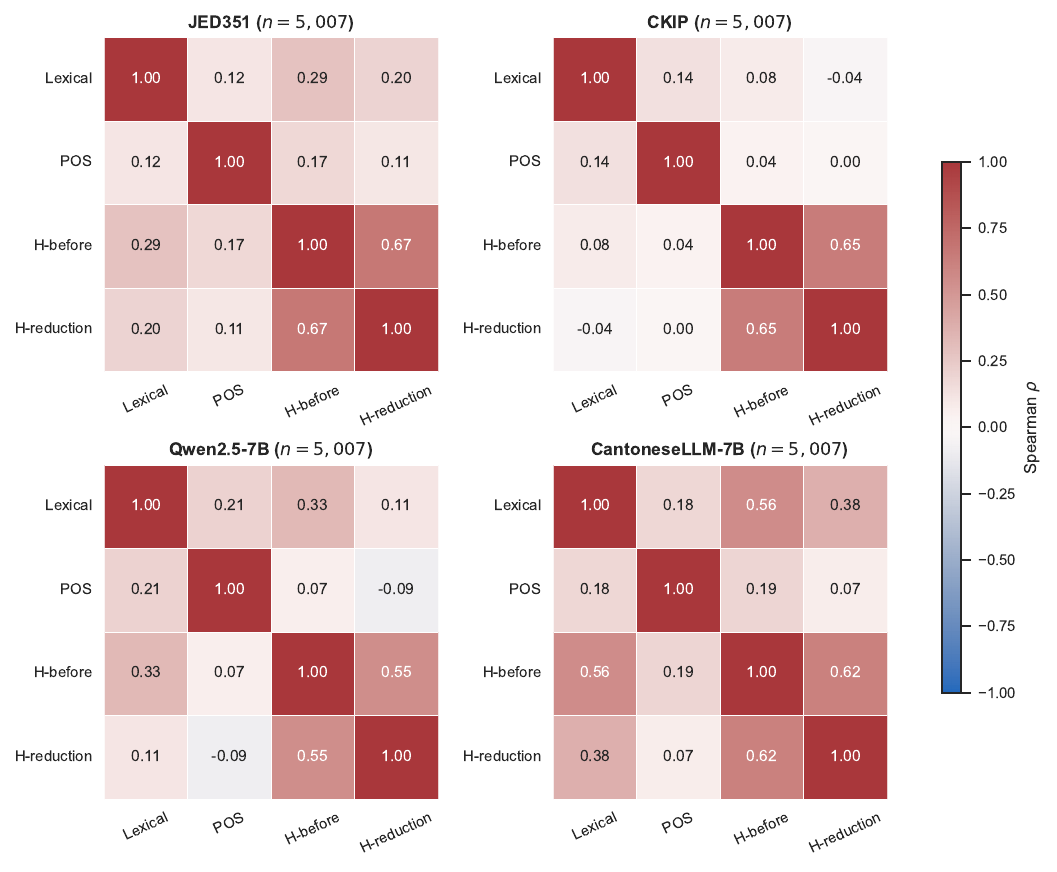}

\caption{Within-model Spearman correlations among the four LLM-derived metrics. Correlations are calculated over 5,007 unique canonical text positions after collapsing the duplicated NR and TSR observations. The full-precision coefficients are provided in the supplementary CSV file.}
\label{fig:metric-correlations}
\end{figure*}

\end{document}